\documentclass[11pt,a4paper]{article}
\usepackage[hyperref]{emnlp2018}
\usepackage{times}
\usepackage{latexsym}

\usepackage{url}

\usepackage{graphicx}
\usepackage{subcaption}

\aclfinalcopy 

\title{Central Yup'ik and Machine Translation of\\Low-Resource Polysynthetic Languages}

\author{Christopher Liu\textsuperscript{1}\footnotemark[1] \quad Laura Domin\'e\textsuperscript{1} \quad Kevin Chavez\textsuperscript{1} \quad Richard Socher\textsuperscript{2} \\
  \textsuperscript{1}Stanford University, Stanford, CA 94305, USA \\
  \textsuperscript{2}Salesforce Research \\
  {\tt \{cwtliu, ldomine, kechavez\}@stanford.edu, rsocher@salesforce.com} 
}

\date{}

\begin{document}

\maketitle

\renewcommand*{\thefootnote}{\fnsymbol{footnote}}
\footnotetext[1]{Corresponding author.}
\renewcommand*{\thefootnote}{\arabic{footnote}}

\begin{abstract}
Machine translation tools do not yet exist for the Yup'ik language, a polysynthetic language spoken by around 8,000 people who live primarily in Southwest Alaska. We compiled a parallel text corpus for Yup'ik and English and developed a morphological parser for Yup'ik based on grammar rules. We trained a seq2seq neural machine translation model with attention to translate Yup'ik input into English. We then compared the influence of different tokenization methods, namely rule-based, unsupervised (byte pair encoding), and unsupervised morphological (Morfessor) parsing, on BLEU score accuracy for Yup'ik to English translation. We find that using tokenized input increases the translation accuracy compared to that of unparsed input. Although overall Morfessor did best with a vocabulary size of 30k, our first experiments show that BPE performed best with a reduced vocabulary size.
\end{abstract}

\section{Introduction}

The Yup'ik language belongs to the Inuit-Yupik-Unangan family whose languages are polysynthetic: their words are made up of many morphemes, yielding a high morphemes-to-word ratio. Thus, a word in Yup'ik can be equivalent to a whole sentence in English. As an example, \textit{pissuryullrunrituk} translates to the English sentence \textit{The two did not want to go hunting}. Only a small amount of parallel texts exist for this historically oral language and no language processing tools have yet been created for Yup'ik. Hence Yup'ik is also a low-resource language which poses unique challenges and trade-offs for reliable machine translation.


Tokenization is usually the first preprocessing step of a machine translation pipeline. Neural networks can only learn a finite number of words in vocabulary and will show poorer performance if the size of the vocabulary is too large. Polysynthetic languages in particular suffer from this issue.

Our primary goal was to train a neural machine translation (NMT) model to reliably translate words from Yup'ik to English. In order to address open vocabulary issues, we segmented Yup'ik words using supervised and unsupervised methods upstream of the NMT model. This paper is organized as follows. Section 2 describes related work. Section 3 introduces the tokenization strategies used for Yup'ik. Sections 4 describes the dataset and neural network architecture. Sections 5 and 6 describe the experimental setup and the resulting comparison of tokenization strategies. Finally, Section 7 presents our conclusions. 

\section{Related work}
The Inuit-Yupik-Unangan language family has been the subject of little to no machine translation research. These languages are agglutinative, necessitating a focus on morphological segmentation as an optimal tokenization strategy for machine translation \cite{vincent}. More recently, unsupervised tokenization schemes have been shown to compete with morphologically pre-processed parsing schemes for machine translation \cite{sennrich2015neural}. Our paper similarly compared unsupervised tokenization NMT performance to that of supervised morphological tokenization.

We chose to use a bidirectional recurrent neural network (RNN) with an attention mechanism according to the built-in seq2seq encoder-decoder framework provided by Tensorflow \cite{luong17}. The seq2seq tutorial included informative use cases, model sizes and complexity chosen for particular corpora sizes. We selected our initial parameters according to these metrics and the agglutinative language type. Finally we added an attention mechanism as a way to incorporate information from the input sentence in the prediction layer \cite{bahdanau2014neural}.

\section{Tokenization}
Using the rule-based parser that we developed, we compared different tokenization schemes upstream of the NMT model: rule-based, unsupervised byte pair encoding (BPE), and unsupervised Morfessor tokenization. 

We also used the word tokenizer functionality of the NLTK toolkit which delimits words based on punctuation for our unparsed dataset \cite{bird}.

\subsection{Rule-based parsing}
A comprehensive Yup'ik-English grammar book was published in 1995 by Steven Jacobson, a trained mathematician, and his Yup'ik wife, Anna Jacobson. It outlines grammatical rules for Yup'ik morphology with high mathematical structure. We developed a rule-based parser using these existing grammar rules \cite{dictionary}.

\subsection{Unsupervised statistical tokenization}
Another tokenization scheme that we included is an unsupervised method called byte pair encoding (BPE), which learns a vocabulary set from the data itself by iteratively merging the most frequent token pairs, beginning with individual characters as tokens. We used the subword-nmt toolkit and learned vocabulary from both languages separately \cite{sennrich2015neural}

\subsection{Unsupervised morphological tokenization}
The Morfessor 2.0 toolkit includes an unsupervised morphological segmentation tool that we  used in our comparison of tokenization schemes. We applied batch training to the Yup'ik data using default parameters \cite{virpioja2013morfessor}.

\begin{figure}[t!]
	\centering
   \label{pipeline} 
   \includegraphics[width=\linewidth]{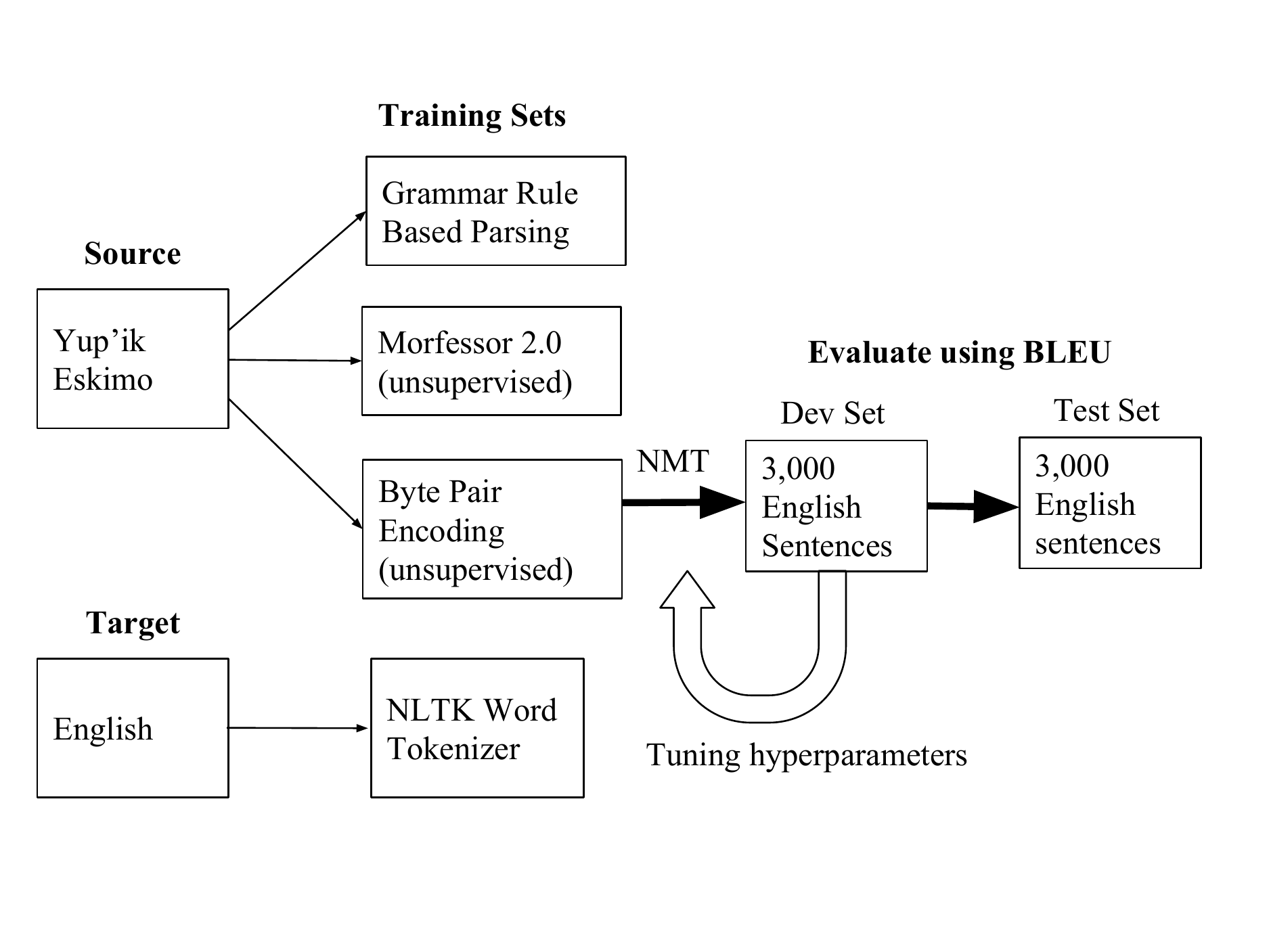}
    \caption{Overview of our project pipeline.}
\end{figure}

\section{NMT setup}
\subsection{Dataset}
Our dataset is made of conversational parallel text in Yup'ik and English from 10 books, totaling roughly 100k lines (averaging 18 English words per line) that were manually scanned with object character recognition. The books were used with permission and are edited by Ann Fienup-Riorden, Alice Rearden, Marie Meade, and others. They contain written transcriptions of interviews with Yup'ik people from various regions across Southwest Alaska, which represents primarily Coastal and Lower Kuskokwim dialects. In addition, a translated copy of the Yup'ik Bible was used which contains mainly narrative text unlike the style of the conversational corpora.
Furthermore, the dataset was divided into train/dev/test (93/3.5/3.5) datasets using 3,500 randomly selected sentences for each of the development and test sets.

\subsection{Architecture}
We used state-of-the-art recurrent neural network (RNN) architectures, specifically bidirectional models with an attention mechanism.

Due to their recent successes, RNNs have become widely used for machine translation applications. Our network has a many-to-many architecture. Due to the low-resource nature of Yup'ik machine translation, the chosen architecture has a shallow depth.

Bidirectional RNNs address a typical shortcoming of RNNs: although they evaluate an input sequence one word at a time, they avoid important contributions from earlier units. Additional backward recurrent units take past and future words into account when making predictions. Long short-term memory (LSTM) units were also used as part of the network architecture following \cite{sutskever2014sequence}. 

Finally an attention mechanism was used in each experiment as these mechanisms have generally received much success in recent years \cite{bahdanau2014neural,luong2015effective}. It is particularly effective in addressing the word order differences between English and Yup'ik translations.

\section{Experimental setup}
\paragraph{Experiments}
Our first set of experiments compared tokenization strategies using 30k-words datasets. Experiment 1 used unparsed Yup'ik as input and was word-tokenized using the NLTK toolkit to delimit punctuation. Experiment 2 used rule-based parsed input, including unresolved words in unparsed form, generated by our implemented parsing method. Experiment 3 used the Morfessor 2.0 toolkit using default batch training to tokenize Yup'ik input \cite{virpioja2013morfessor}. Finally, Experiment 4 used BPE using 30k merge operations to generate input \cite{sennrich2015neural}

The second set of experiments were designed to evaluate translation performance using only unsupervised BPE tokenization. We compared results using merge operation counts of 10k, 15k, and 30k. By comparing these BPE parsed datasets with varying vocabulary size, we found that 15k merges returned highest performance for this dataset.

\paragraph{Hyperparameter tuning}
Hyperparameters were carefully considered in order to make conclusive arguments about optimal tokenization strategy.
After hyperparameter search, each model was run with the following parameters: learning rate (0.5), exponential learning rate decay, number of layers (2), number of steps (80k), maximum sequence length (50), number of units (128), and batch size (128).

\paragraph{Evaluation method}
The bilingual evaluation study (BLEU) is a standard accuracy metric for machine translation \cite{papineni2002bleu}. The BLEU implementation we used from Moses \cite{koehn2007moses} returns the geometric average of n-gram BLEU scores and multiplies that result by an exponential brevity penalty factor.

\[
BLEU = BP \large( \sum_{n=1}^N \frac{\log P_n}{N} \large)
\]

Here $P_n$ denotes the precision of n-grams in the hypothesis translation, and the $N$ we chose was $4$.

\section{Results}

\begin{table}[t]
\label{table1}
\begin{center}
\begin{tabular}{llll}
\multicolumn{1}{c}{\bf Exp \#}  &\multicolumn{1}{c}{\bf Yup'ik Source} &\multicolumn{1}{c}{\bf Dev} &\multicolumn{1}{c}{\bf Test}
\\ \hline \\
1         &Unparsed NLTK        &9.58         &9.02 \\
2         &Rule-based         &8.51         &8.33 \\
3         &Morfessor        &13.33         &12.59 \\
4         &BPE 30k merges         &12.39         &11.77 \\
5         &BPE 15k merges         &{\bf 13.52}         &{\bf 12.71} \\
6         &BPE 10k merges         &13.19         &12.66 \\
\end{tabular}
\end{center}
\caption{BLEU scores at step with highest score (out of 100). Dev stands for development dataset and test stands for test dataset.}
\end{table}

\subsection{Tokenization strategy (experiments 1-4)}
Both Morfessor and BPE 30k (unsupervised) datasets outperformed the baseline unparsed NLTK dataset. These results suggest that, after controlling for vocabulary size, parsed methods are preferable to unparsed methods for increased prediction accuracy.

Contrary to our initial hypothesis, the rule-based parsed dataset performed worse than the baseline unparsed method. This is likely due to the remaining unresolved Yup’ik words that did not match to a set of morphemes in our parser, or to out-of-vocabulary issues.

\begin{figure}[t!]
        \centering
        \includegraphics[height=1.7in]{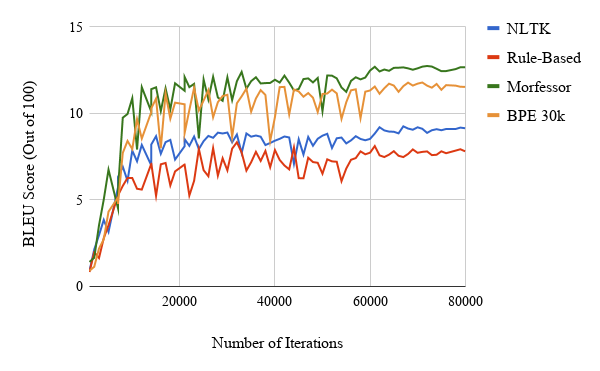}
        \caption{Test BLEU for each tokenization method (30k vocab size) on the test set.}
\end{figure}

\begin{figure}
        \centering
        \includegraphics[height=1.7in]{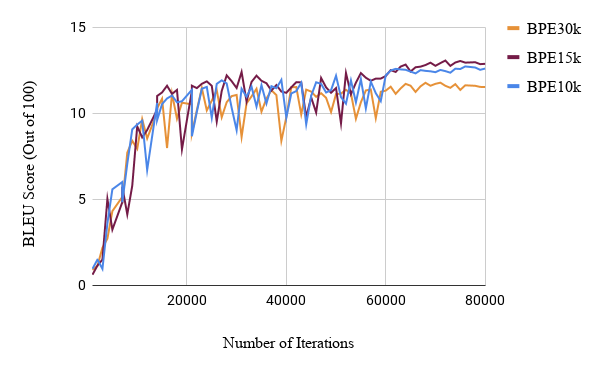}
        \caption{Test BLEU for each BPE method with varying vocabulary size on the test set.}
\end{figure}

\begin{table*}[h]
\begin{center}
\begin{tabular}{ll}
\multicolumn{1}{c}{\bf Tokenization}  &\multicolumn{1}{c}{\bf Prediction Sentence 1} 
\\ \hline \\
\textbf{Original}         &\textit{\textbf{in the spring, they hauled seals up from the ocean. }}         \\
Unparsed         &\textit{in spring , spotted seals always spotted seals . }            \\
Rule-based         &\textit{in spring they had a seal spotted seal oil , and they were unable to catch them . }             \\
Morfessor         &\textit{in the spring , they brought spotted seals up there , they 'd haul spotted seals up there . }             \\
BPE 30k         &\textit{in the spring they were bringing supplies to the ocean in the ocean . }             \\
BPE 15k         &\textit{in spring they brought the spotted seals up in the spring . }             \\
BPE 10k         &\textit{in the spring , they brought grass up to shore . }             \\
\end{tabular}
\end{center}
\caption{Example sentence predictions for the different tokenization experiments.}
\label{table3}
\end{table*}

\subsection{BPE Merge Count (experiments 4-6)}
Experiment 5 parsed with BPE using 15k merge operations outperformed the other BPE and Morfessor experiments. Vocabulary size is an adjustable parameter that should target optimal parsing levels for Yup'ik. For our Yup'ik dataset, which includes mixing of conversational, narrative, and dialectical differences in domains, unsupervised tokenization returned the highest accuracy for machine translation.

\subsection{Error analysis}
Table \ref{table3} and annex \ref{annex1} contain selected predicted translations using the trained models from all experiments. The models often returned words that were different, yet similar in meaning, compared to the reference translation. In Prediction Sentence 1, Morfessor and BPE 15k are closest in conveying the correct translation. Although our experimental BLEU scores range from 7 to 13 percents in accuracy, the quality of the translations is fairly sufficient in conveying original sentence information. In addition, the particular experiment type that most closely matched reference translation generally varied across sentences.

\section{Conclusion}
Contrary to our expectations about morphology, the tokenization scheme that scored best was not the rule-based parser but an unsupervised tokenization scheme, Morfessor 2.0 toolkit. Our second run of experiments, with BPE parsing only, points to a targeted vocabulary size that might be best suited for Yup'ik. This could be useful for any future studies of this particular language.

Future work could include applying BPE to the rule-based parsed dataset in order to decrease the vocabulary size yielded by the rule-based strategy while retaining its morphological relevance. Another direction worth exploring to improve the translation predictions is postprocessing methodology such as beam search and upgrading to more complex models. Ultimately we plan to reverse the language pair in order to perform bidirectional translation. To rule out some unnatural predictions we could use sentiment prediction and combine it with the NMT model prediction to return the translation with the highest subjective score. Finally, we are working on a phone application that will integrate tools from this project (dictionary lookup, morphological, analysis, and translation between English and Yup'ik) and make it available to the public. Our hope is that this translation tool could fight back language endangerment by facilitating access to translation for Yup'ik language learners.

\paragraph{Source code}
The source code for the Yup'ik parser and the experiments discussed in this paper is available at (link omitted for anonymous submission). The Yup'ik-English parallel corpus is available by request. 


\bibliography{emnlp2018}
\bibliographystyle{acl_natbib_nourl}

\clearpage
\appendix
\section{Examples of translated sentence pairs}
\label{annex1}
Tables 3, 4 and 5 give additional examples of translated sentence pairs for the different tokenization methods. Table 3 contains the word qaygi which means traditional men's community house, a location where hunting was often taught.
\bigskip
\begin{table}[ht]
\centering
\begin{tabular}{ll}
\multicolumn{1}{c}{\bf Tokenization}  &\multicolumn{1}{c}{\bf \hspace{5.5cm}  Prediction Sentence 2 \hspace{4.5cm} } 
\\ \hline \\
 \textbf{Original}         & \textit{\textbf{the dog went hunting and brought food for her .}}         \\
 Unparsed         &\textit{the dog went out to the qaygi to hunt .}       \\
 Rule-based         &\textit{since the dog had n't eaten , he went hunting and hunted with food .}             \\
 Morfessor        &\textit{his dog would try to obtain his dog to obtain food .}             \\
 BPE 30k         &\textit{that one went to dog food for her .}             \\
 BPE 15k         &\textit{his dog started to hunt for food to hunt .}             \\
 BPE 10k         &\textit{the dog had to use the dog for a long time .}         \\\\
\hspace{1cm} & \hspace{1.5cm} {\small Table 3:} Example sentence predictions
\end{tabular}
\label{table5}
\end{table}

\begin{table}[h]
\begin{center}
\begin{tabular}{ll}
\multicolumn{1}{c}{\bf Tokenization}  &\multicolumn{1}{c}{\bf \hspace{5.5cm}  Prediction Sentence 3 \hspace{4.5cm} }
\\ \hline \\
\textbf{Original}         &\textit{\textbf{i know you are not a good hunter .}}        \\
Unparsed         &\textit{i believe you know you are always alone .}         \\
Rule-based         &\textit{i know you , you poor things .}            \\
Morfessor         &\textit{you are in the future .}             \\
BPE 30k         &\textit{i know you , i am a successful hunter .}              \\
BPE 15k         &\textit{i know you were a good hunter . ''}             \\
BPE 10k         &\textit{i know you , i know you are hunting .}             \\ \\
\hspace{1cm} & \hspace{1.5cm} {\small Table 4:} Example sentence predictions (continued)
\end{tabular}
\label{table2}
\end{center}
\end{table}

\begin{table}[h]
\begin{tabular}{ll}
\multicolumn{1}{c}{\bf Tokenization}  &\multicolumn{1}{c}{\bf \hspace{5.5cm}  Prediction Sentence 4 \hspace{4.5cm} }
\\ \hline \\
\textbf{Original}                &\textit{\textbf{when people were roughhousing in their homes, ghosts would appear.}} \\
Unparsed                  &\textit{and when we 'd pick berries , we 'd go and get them to our destination .} \\
Rule-based                    &\textit{and when she got to the intent , she saw that it was a ghost .}\\
Morfessor                 &\textit{and when they 'd constantly have a situation too much , a ghost would surface it too much .} \\
BPE 30k                  &\textit{and when they were talking to us in a community , they would come to the land .} \\
BPE 15k                &\textit{and when they discussed something , ghosts would also pull them up to the ghosts .} \\
BPE 10k                  &\textit{and when they were about to get too rambunctious , they 'd bring a ghost up .}\\ \\
\hspace{1cm} & \hspace{1.5cm} {\small Table 5:} Example sentence predictions (continued)
\end{tabular}
\label{table4}
\end{table}

\end{document}